\documentclass[10pt,journal]{IEEEtran}

\ifCLASSINFOpdf
\else
\fi
\usepackage{times}
\usepackage{epsfig}
\usepackage{graphicx}
\usepackage{amsmath}
\usepackage{amssymb}

\usepackage{footnote}
\usepackage{multirow}
\usepackage{booktabs}
\usepackage{threeparttable}
\usepackage{enumerate}
\usepackage{indentfirst}
\usepackage{array}
\usepackage{rotating}
\usepackage{ragged2e}
\usepackage{wasysym}
\usepackage{caption}
\usepackage{epstopdf}
\usepackage{color}

\def\ie{{\em i.e.}}
\def\eg{{\em e.g.}}
\def\etal{{\em et al.}}

\newcommand{\figref}[1]{Fig. \ref{#1}}
\newcommand{\tabref}[1]{Tab. \ref{#1}}

\newcommand{\secref}[1]{Section \ref{#1}}

\newcommand{\bl}[1]{\textbf{#1}}
\newcommand{\mc}[1]{\mathcal{#1}}
\newcommand{\mb}[1]{\mathbb{#1}}

\newcommand{\bm}[1]{\mbox{\boldmath{$#1$}}}
\begin{document}
%
\title{Model-guided Multi-path Knowledge Aggregation for Aerial Saliency Prediction}
%
%
%

\author{Kui~Fu,
        Jia~Li,~\IEEEmembership{Senior Member,~IEEE},
        Yu Zhang,~\IEEEmembership{Member,~IEEE},\\
        Hongze Shen and~Yonghong Tian,~\IEEEmembership{Senior Member,~IEEE}
\IEEEcompsocitemizethanks{\IEEEcompsocthanksitem Kui Fu is with the State Key Laboratory of Virtual Reality Technology and Systems, School of Computer Science and Engineering, Beihang University, Beijing 100191, China, and also with the Peng Cheng Laboratory, Shenzhen 518000, China. \protect
\IEEEcompsocthanksitem Jia Li is with the State Key Laboratory of Virtual Reality Technology and Systems, School of Computer Science and Engineering, Beihang University,
Beijing 100191, China, also with the Beijing Advanced Innovation Center for Big Data and Brain Computing, Beihang University, Beijing 100191, China,
and also with the Peng Cheng Laboratory, Shenzhen 518000, China (e-mail: jiali@buaa.edu.cn). \protect
\IEEEcompsocthanksitem Yu Zhang and Hongze Shen are with the State Key Laboratory of Virtual Reality Technology and Systems, School of Computer Science and Engineering, Beihang University, Beijing 100191, China. \protect
\IEEEcompsocthanksitem Yonghong Tian is with the National Engineering Laboratory for Video Technology, Department of Computer Science and Technology, Peking University, Beijing 100871, China, and also with the Peng Cheng Laboratory, Shenzhen 518000, China. \protect
\IEEEcompsocthanksitem J. Li is the corresponding author. URL: http://cvteam.net. \protect
}
}

%
%

\markboth{ }%
{Shell \MakeLowercase{\textit{et al.}}: Bare Demo of IEEEtran.cls for IEEE Journals}
%



\maketitle

\begin{abstract}
  As an emerging vision platform, a drone can look from many abnormal viewpoints which brings many new challenges into the classic vision task of video saliency prediction. To investigate these challenges, this paper proposes a large-scale video dataset for aerial saliency prediction, which consists of ground-truth salient object regions of 1,000 aerial videos, annotated by 24 subjects. To the best of our knowledge, it is the first large-scale video dataset that focuses on visual saliency prediction on drones. Based on this dataset, we propose a Model-guided Multi-path Network (MM-Net) that serves as a baseline model for aerial video saliency prediction. Inspired by the annotation process in eye-tracking experiments, MM-Net adopts multiple information paths, each of which is initialized under the guidance of a classic saliency model. After that, the visual saliency knowledge encoded in the most representative paths is selected and aggregated to improve the capability of MM-Net in predicting spatial saliency in aerial scenarios. Finally, these spatial predictions are adaptively combined with the temporal saliency predictions via a spatiotemporal optimization algorithm. Experimental results show that MM-Net outperforms ten state-of-the-art models in predicting aerial video saliency.
\end{abstract}


%
\IEEEpeerreviewmaketitle

\section{Introduction}\label{sec:intro}

\IEEEPARstart{V}{isual} saliency prediction is one of the fundamental vision problems that has been extensively studied for several decades \cite{itti1998model, Borji2012State,Nguyen2017Attentive}. With the proposal of comprehensive rules \cite{Goferman2012Context, Li2015Finding}, large training datasets \cite{deng2009imagenet, CAT2000, abu2016youtube} and deep learning algorithms \cite{Pan2016Shallow, Fang2017Learning, zhang2019learning, zhang2019cousin}, the performance of saliency models has been improving steadily. Meanwhile, many saliency-based attentive systems have achieved impressive performance in image recognition \cite{Zhou2014Learning}, video compression \cite{Gupta2013Visual}, content-based adverting \cite{Shen2014Webpage}, robot interaction \cite{Muhl2007On} and navigation \cite{Gage2012Saliency}. Despite the success of various existing models, an important concern still remains in the literature: whether visual knowledge extracted from existing saliency models can boost the performance of saliency prediction in various scenarios, especially the newly emerging ones?

In most existing works, this question is explored using plenty of images \cite{CAT2000} and videos \cite{Fang2014An} collected from Internet to cover various daily scenarios taken by digit cameras and mobile phones. In this paper, we conduct research on an emerging but less studied domain, the aerial videos captured by drones. A drone can observe the world from many different viewpoints, providing us an opportunity to revisit the problem from new perspectives.
In particular, from these aerial videos, we wish to explore reliable answers to two questions in visual saliency estimation:

\noindent 1.~Whether previous ground-level saliency models still work well in processing aerial videos?

\noindent 2.~How to transfer the ground-level knowledge to aerial platforms to develop an aerial saliency model?

To answer these two questions, we collect 1,000 videos captured by drones, which are then free-viewed by 24 subjects in eye-tracking experiments to collect dense, accurate fixations. In this manner, the ground-truth salient regions in aerial videos can be well annotated, based on which we can benchmark the performance of various visual saliency models deployed on drones. By testing the performance of ten ground-level saliency models, we find these models still work impressively in capturing salient regions in aerial videos. However, there still exist large gap between their predictions and the ground-truth maps. Therefore, it is necessary to explore the way that drones look and develop a saliency model suitable for the saliency prediction task on aerial videos.

\begin{figure}[t]
	\begin{center}
		\includegraphics[width=1.0\columnwidth]{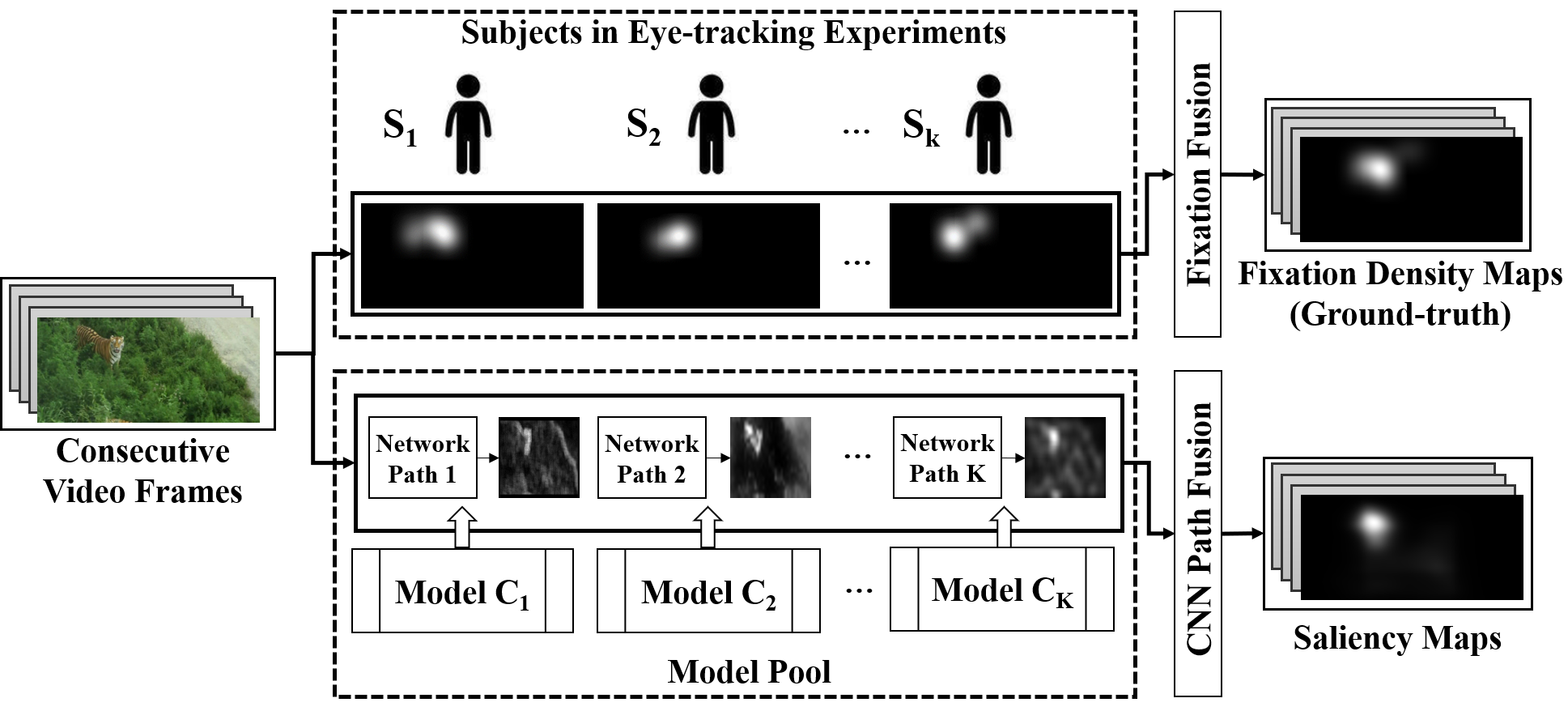}
	\end{center}
	\caption{In eye-tracking experiments, saliency maps can be generated by fusing fixations of multiple subjects. This process motivates the design of a multi-path network architecture, in which different paths are guided by different classic saliency models to encode different knowledge about visual saliency prediction. }
	\label{fig:motivation}
\end{figure}

\begin{figure*}[t]
	\begin{center}
		\includegraphics[width=1.0\textwidth]{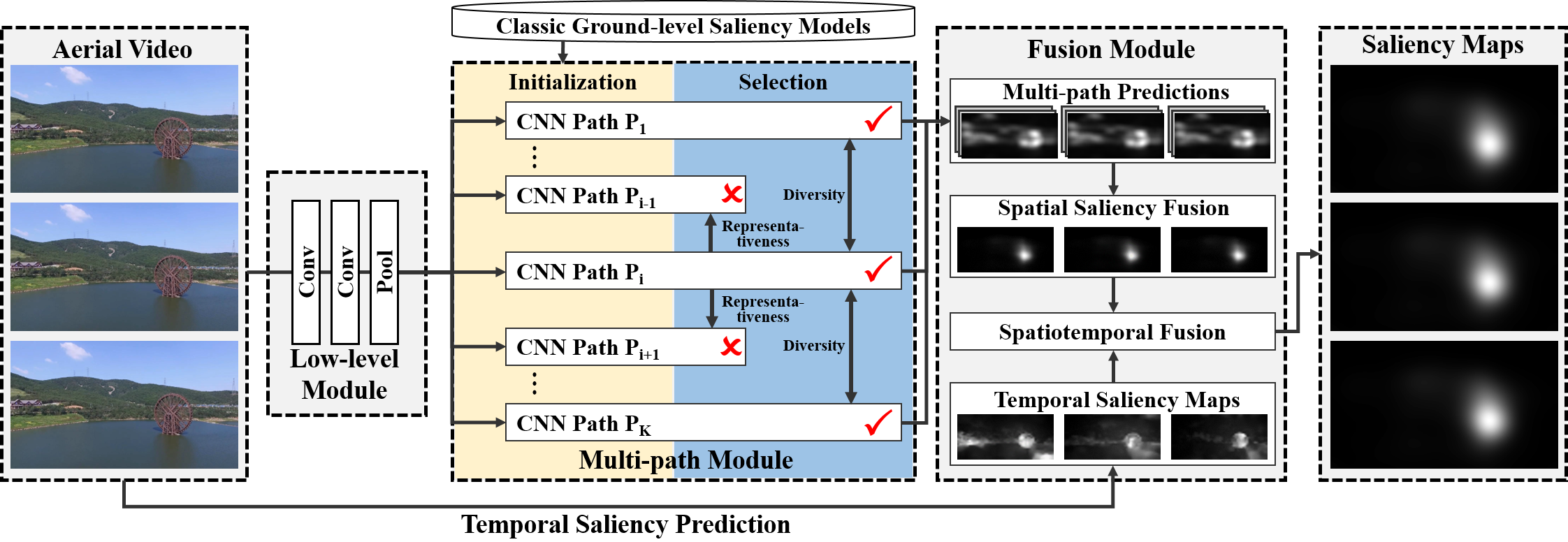}
	\end{center}
	\caption{System framework of our proposed baseline model MM-Net. The low-level module extracts low-level features. After that, these features are delivered into a multi-path module, in which each path is a sub-network pre-trained on massive aerial scenes under the guidance of a classic saliency model. In this manner, various visual saliency knowledge can be encoded in multiple paths to enhance the capability of MM-Net in processing aerial videos. To reduce computational cost, we further propose a selection algorithm to remove redundant paths according to path representativeness, diversity as well as module complexity. After that, the selected paths are fine-tuned on aerial scenarios to generate spatial saliency maps. They are then adaptively fused with the temporal saliency predictions to obtain clean and accurate saliency prediction results.}
	\label{fig:framework}
\end{figure*}

Based on this dataset, we also develop a Model-guided Multi-path Network (MM-Net) that can be used as a baseline model for aerial video saliency prediction. The design of MM-Net is motivated by the settings of eye-tracking experiments (see Fig.~\ref{fig:motivation}) and aims to aggregate the visual saliency knowledge from many classic models into a unified deep network. As shown in \figref{fig:framework}, common low-level features are first extracted in a low-level module of MM-Net, which are then fed into a multi-path module. This module contains multiple information paths, each of which is initialized under the guidance of a classic saliency model. After that, redundant paths are identified and removed via a path selection algorithm that jointly considers path diversity, representativeness and the overall complexity of the multi-path module. The selected paths then enter the fusion module, and the visual saliency knowledge encoded in the selected paths are aggregated to predict spatial saliency \cite{ding2018improving} in aerial scenarios. Finally, spatial saliency maps can be efficiently predicted, which are then adaptively fused with the temporal saliency predictions to obtain clean and accurate saliency maps for various types of aerial videos. Experimental results show that MM-Net outperforms ten state-of-the-art models on aerial videos.

The main contributions of this paper include: 1)~We propose, to the best of our knowledge, the first large-scale video dataset for aerial saliency prediction; 2)~We propose a Model-guided Multi-path Network that provides a way to transfer the knowledge from multiple classic models into a single deep model; 3)~We propose an effective path selection algorithm, which can be used to balance the complexity and effectiveness of the multi-path network.

The rest of this paper is organized as follows: \secref{sec:related works} reviews related works and \secref{sec:dataset} presents the dataset. Our approach is presented in \secref{sec:methods} and tested in \secref{sec:experiments}. Finally, \secref{sec:conclusion} concludes the paper.

\section{Related Works}\label{sec:related works}
In this section, we present a brief review of computational saliency models from three aspects: heuristic models, non-deep learning models and deep learning models.

\textbf{Heuristic saliency models} can be roughly categorized into bottom-up \cite{itti1998model, seo2009static} and top-down categories \cite{jiang2013salient, li2014saliency}. The bottom-up models are stimulus-driven and infer saliency from visual stimuli themselves with hand-crafted features (\eg,~direction, color and intensity) and/or limited human knowledge (\eg,~center-bias). Due to the imperfect hand-crafted features or heuristic fusion strategies, bottom-up models may have some difficulties in suppressing background distractors. To address this problem, some top-down models heuristically incorporate high-level factors. For example, Borji \etal~\cite{borji2012probabilistic} proposed an unified Bayesian approach to integrate global context of a scene, previous attended locations and previous motor actions over time to predict the next attending locations. Chen \etal~\cite{chen2016video} proposed a video saliency model that predicted video saliency by combining the top-down saliency maps with the bottom-up ones through point-wise multiplication.

\textbf{Non-deep learning saliency models} propose to learn the fusion strategies of various heuristic saliency cues~\cite{li2010probabilistic, vig2012intrinsic, Vig2014Large}. For example, Vig \etal~\cite{vig2012intrinsic} proposed a simple bottom-up model for dynamic scenarios with the aim of keeping the number of salient regions to a minimum. Recently, Fang \etal~\cite{Fang2017Learning} proposed an image saliency model by learning a set of discriminative subspaces that perform the best in popping out targets and suppressing distractors. Li \etal~\cite{li2016measuring} proposed a saliency model that measured the joint visual surprise from intrinsic and extrinsic contexts. However, the hand-crafted features used in these models may inherently set an upper bound for the final performance.

\textbf{Deep learning saliency models} have great advantages in learning feature representations~\cite{hu2017deep, kuen2016recurrent, li2016deepsaliency}. For example, K\"{u}mmerer \etal~\cite{kummerer2014deep} presented a Convolutional Neural Network (CNN) that reused AlexNet \cite{krizhevsky2012imagenet} to generate high-dimensional features. Pan \etal~\cite{Pan2016Shallow} proposed two designs, a shallow network and a deeper network, that can be trained end-to-end for fixation prediction. Lahiri \etal~\cite{lahiri2016wepsam} proposed a saliency model which used a two step learning strategy.
These deep learning models usually have high computational efficiency and impressive performance in daily scenarios. However, it is unclear whether these saliency models can be reused in aerial platforms, which has remarkable viewpoint and depth changes. Therefore, it is necessary to construct a large-scale aerial saliency dataset to benchmark saliency models.

\section{The Aerial Video Saliency Dataset} \label{sec:dataset}
In this section, we follow \cite{li2018benchmark} to label the salient regions in the videos by eye-tracking and present a large-scale dataset for aerial video saliency. Note that such an annotation process is a regular process in the field of visual video attention prediction and minimizes the bias from different subjects via a voting manner, leading to accurate annotations. We also benchmark classic ground-level models to show the difference and correlation between aerial and ground-level saliency prediction.
\subsection{Dataset Construction}
To construct the dataset, we download hundreds of long aerial videos from Internet that are captured by drones. We manually divide these long videos into shots and randomly sample 1,000 shots with a total length of 1.6 hours (\ie, 177,664 frames at 30 FPS). We find that the dataset mainly covers videos from four genres: building, human, vehicle, and others. Thus the dataset, denoted as AVS1K\footnote{Dataset is publicly available at http://cvteam.net}, contains four subsets that are denoted as {AVS1K-B}, {AVS1K-H}, {AVS1K-V} and {AVS1K-O}, respectively.

To annotate the ground-truth salient regions, we conduct massive eye-tracking experiments involved with 24 subjects. For each video in the dataset, we collect eye fixations on each frame and convert them to fixation density maps following the setting of \cite{li2018benchmark}, making the whole dataset densely annotated. Note that each video is free-viewed by 17-20 randomly selected subjects. All subjects have normal or corrected to normal vision, and they have never seen these videos before. Note that we do not provide any prior information to the subjects and let them watch videos in a free-viewing manner in the eye-tracking experiments. In experiments, the videos are displayed on a 22-inch color monitor with the resolution of $1680\times 1050$. A chin set is adopted to eliminate the error caused by the head wobble and fix the monitor viewing distance to 75cm. Other experimental conditions such as illumination and noise are set to constant for all subjects.

\begin{figure}[t]
\begin{center}
	\includegraphics[width=1.0\columnwidth]{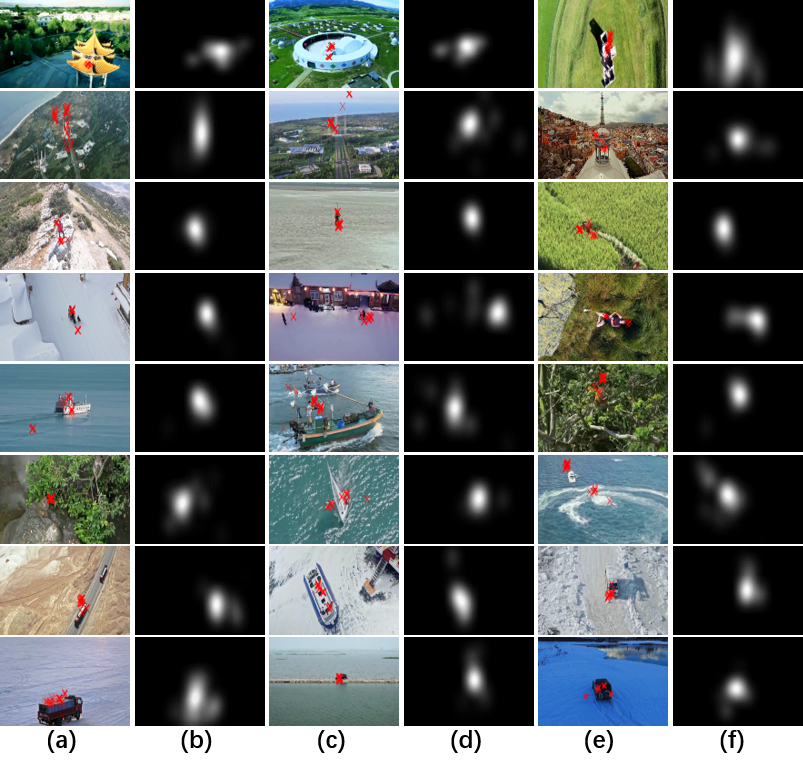}
\end{center}
\caption{Sample aerial frames (a, c, e), fixations (red dots) and ground-truth saliency maps (b, d, f) from AVS1K.}
\label{fig:ground-truth maps}
\end{figure}

Given the fixation data, we can compute a fixation density map for each frame to annotate the ground-truth salient regions, \ie, the salient regions that a drone should look at from the perspective of human-being. Let $\mc{I}_t\in\mc{V}$ be a frame presented at time $t$, we measure the fixation density map of $\mc{I}_t$, denoted as $S_t$ as in \cite{li2018benchmark}. The value of $S_t$ at pixel $p$ can be computed as
\begin{equation}
\label{eq:1}
S_t(p) = \sum_{f\in\mb{F}_\mc{V}}\delta(t_f\geq{}t)\cdot{}D_{spa}(f,p)\cdot{}D_{tem}(f,p),
\end{equation}
where $D_{spa}(f, p)$ and $D_{tem}(f,p)$ measure the spatial and temporal influences of the fixation $f$ to the pixel $p$, respectively. By using an indicator function $\delta(t_f\geq{}t)$ that equals 1 if $t_f\geq{}t$ and 0 otherwise, we only consider the influence of fixations in a short period after $t$. Let $(x_p, y_p)$ be the coordinate of $p$, the values of $D_{spa}(f, p)$ and $D_{tem}(f,p)$ can be computed as
\begin{equation}
\begin{split}
D_{spa}(f,p)=&\exp{\left(-\frac{(x_f-x_p)^2+(y_f-y_p)^2}{2\sigma^2_D}\right)},\\
D_{tem}(f,p)=&\exp{\left(-\frac{(t_f-t)^2}{2\sigma^2_T}\right)},
\end{split}
\label{eq:influence}
\end{equation}
where $\sigma_D$ and $\sigma_T$ are two constants to control the spatial and temporal influences of fixations, which are empirically set to 3\% of video width (or video height if it is larger) and 0.1s, respectively. Some representative frames, recorded fixations and generated ground-truth saliency maps can be found in \figref{fig:ground-truth maps}. Dataset statistics can be found in \tabref{tab:detail_AVS1K}.

\subsection{Model Performance in Aerial Scenarios}
As it stands, aerial videos often have higher viewpoints, a wider field of vision and smaller targets. In other words, the visual patterns in aerial videos may be remarkably different from those on the ground. Thus, it is worth exploring the performance of classic saliency models in aerial scenarios. To address this concern, we test ten classic saliency models on AVS1K. These models include AIM \cite{bruce2007attention}, AWS \cite{garcia2012relationship}, BMS \cite{zhang2013saliency}, GB \cite{harel2007graph}, HFT \cite{li2013visual}, ICL \cite{hou2009dynamic}, IT \cite{itti1998model}, QDCT \cite{schauerte2012quaternion}, SP \cite{li2014visual} and SUN \cite{zhang2008sun}. Note that they are not learning-based and thus less sensitive to dataset bias. As an intuitive comparisons, we also show the performance of these models over DHF1K~\cite{wang2018revisiting}, the latest large-scale video saliency dataset fulfilled with daily videos collected by digital cameras and mobile phones.

\begin{table}[t]
\small
\centering
\caption{Dataset statistics of the AVS1K dataset.}
\label{tab:detail_AVS1K}
\begin{tabular}{l|cccc}
	\toprule
	~Dataset               &~Video &~Max Res.        &~Frames &~Avg. Len. (s)  \tabularnewline
	\midrule
	{AVS1K-B}              &~240   &~1280$\times$720 &~41,471 &~5.76           \tabularnewline
	{AVS1K-H}              &~210   &~1280$\times$720 &~31,699 &~5.03           \tabularnewline
	{AVS1K-V}              &~200   &~1280$\times$720 &~27,092 &~4.52           \tabularnewline
	{AVS1K-O}              &~350   &~1280$\times$720 &~77,402 &~7.37           \tabularnewline
	AVS1K                  &~1000  &~1280$\times$720 &~177,664&~5.92           \tabularnewline
	\bottomrule
\end{tabular}
\end{table}
\begin{table}[t]
	\footnotesize
	\centering{
		\caption{Performance comparisons on AVS1K and DHF1K.} \label{tab:heuristic}
		\begin{tabular}{l@{}|ccc|ccc}
			\toprule
			\multirow{2}*{Method~}&\multicolumn{3}{c|}{AVS1K}&\multicolumn{3}{c}{DHF1K} \tabularnewline
			&~AUC&~sAUC&~NSS&~AUC&~sAUC&~NSS\tabularnewline \midrule
			AIM  &~0.644 &~0.652 &~0.944 &~0.713 &~0.576 &~0.812 \tabularnewline
			AWS  &~0.681 &~0.728 &~1.391 &~0.693 &~0.571 &~0.854 \tabularnewline
			BMS  &~0.710 &~0.746 &~1.660 &~0.743 &~0.580 &~1.031 \tabularnewline
			GB   &~0.612 &~0.650 &~0.984 &~0.631 &~0.558 &~0.599 \tabularnewline
			HFT  &~0.789 &~0.715 &~1.671 &~0.806 &~0.579 &~1.323 \tabularnewline
			ICL  &~0.698 &~0.651 &~1.252 &~0.712 &~0.535 &~0.697 \tabularnewline
			IT   &~0.540 &~0.572 &~0.844 &~0.559 &~0.517 &~0.387 \tabularnewline
			QDCT &~0.689 &~0.696 &~1.302 &~0.720 &~0.569 &~0.909 \tabularnewline
			SP   &~0.781 &~0.706 &~1.602 &~0.811 &~0.581 &~1.400 \tabularnewline
			SUN  &~0.587 &~0.603 &~0.672 &~0.628 &~0.546 &~0.540 \tabularnewline
			\bottomrule
		\end{tabular}
	}
\end{table}

To measure the performance of these models, we select three widely adopted metrics according to~ \cite{riche2013saliency,li2015data,bylinskii2016different}, including the traditional Area Under the ROC Curve (AUC), the shuffled AUC (sAUC) and the Normalized Scanpath Saliency (NSS). Typically, AUC may assign high scores to a fuzzy saliency map if it correctly predicts the orders of salient and less-salient locations, while sAUC and NSS prefer clean saliency maps that only pop-out the most salient locations and suppress all the other regions.

Based on the three metrics, \tabref{tab:heuristic} shows the performance of ten classic models on AVS1K and DHF1K. From \tabref{tab:heuristic}, we find that the AUC scores of all models on AVS1K are lower than those on DHF1K. Such inferior performance intuitively demonstrates that it is challenging for classic saliency models in dealing with aerial scenarios since they are designed based on human visual mechanisms on the ground-level but can not directly fit to aerial scenarios. Note that it is not affected by the bias of ground-truths labeling. This may be caused by the fact that aerial videos often have wider field of vision and thus contain richer contents. Surprisingly, sAUC and NSS, which focus on the saliency amplitude, achieve even higher scores on AVS1K than on DHF1K. This implies that the salient targets in aerial videos, which are usually very small, demonstrate impressive capability to pop-out from its surroundings from the higher viewpoints. These results imply that visual saliency knowledge encoded in classic models can be reused in aerial saliency prediction after certain domain adaptation operations.

\section{The Model-guided Multi-path Network}\label{sec:methods}

\subsection{Path Initialization}
Motivated by above observations, we present MM-Net, a baseline model that absorbs the visual saliency knowledge in classic models and evolves to handle aerial video saliency like the human being does. As shown in Fig.~\ref{fig:framework},  MM-Net starts with a low-level module that consists of two convolution layers and one max pooling layer. We initialize the parameters of the low-level module with the first two convolution layers of VGG16~\cite{simonyan2014very}. Given these low-level features, there exist many ways in classic models to extract and fuse saliency cues from them. To make use of the knowledge in these models, we select the ten classic models we have tested in \secref{sec:dataset}, each of which is used to guide the initialization process of a network path in Fig.~\ref{fig:framework}. In the initialization, we first obtain the saliency maps of a classic model on the training set (500 videos) and validation set (250 videos) of AVS1K. These model-estimated saliency maps are then used as ``ground-truth'' to fine-tune the layers in each MM-Net path. Note that the multiple paths and fusion modules are independently trained. The selected three paths are fixed in the fusion process. Each path is initialized with Xavier's algorithm and the input resolution is $320 \times 320$.
In this process,the low-level module is fixed so that the parameters of each network path are independently updated.
By minimizing the cross entropy loss between path outputs and classic model predictions, each path is forced to behave like a classic model so as to distillate its knowledge of saliency prediction.

After initialization, MM-Net inherently learns how to extract and fuse various saliency-related features. However, the knowledge encoded in different paths is highly redundant, and how to remove such redundancy to reduce model complexity is the next issue to be addressed.

\subsection{Path Selection} \label{subsec:path selection}
To remove the path redundancy, we propose a path selection algorithm that jointly considers path diversity, representativeness and the overall complexity of the multi-path module. Formally, a binary column vector $\bm{\alpha}$  with $M$ binary components ($M=10$ in this study) is adopted, which indicates the $i$th path is selected if $\bm{\alpha}_i = 1$, or discarded otherwise. For all the $K$ frames from the $250$ validation videos of AVS1K, we denote the saliency map predicted by the $i$th path on the $k$th frame as $S_{k_i}$. As a result, the path selection process can be solved by optimizing
\begin{equation}\label{eq:optobj}
\begin{split}
\bm{\alpha}^*=&\arg\max_{\bm{\alpha}}~\Omega_r+\lambda_d\Omega_d\\
\text{s.t.}~&1\leq{}\|\bm{\alpha}\|_0\leq{}M,~\text{and}~\alpha_i\in\{0,1\},\forall{}~i
\end{split}
\end{equation}
where $\|\bm{\alpha}\|_0$ denotes the number of non-zero components in $\bm{\alpha}$ and thus reflects the complexity of the multi-path module. The terms $\Omega_r$ and $\Omega_d$ denote the representativeness and diversity to be maximized, respectively. The $\lambda_d$ is a weight parameter to balance the representativeness and diversity, which is empirically set to 0.2 (its influence on final results will be discussed in experiments).

The term $\Omega_r$ is defined according to path similarities. That is, the unselected paths should be highly similar to selected ones that are considered to be representative. This term can be defined as
\begin{equation}\label{eq:penaltyRep}
\Omega_{r}=\frac{\sum\limits_{i=1}^{M}(1-\alpha_i)\cdot{}\max\{\alpha_j\cdot\text{Sim}_{ij}|\forall~j\neq{}i\}}{\sum\limits_{i=1}^{M}(1-\alpha_i)+\epsilon},
\end{equation}
where $\epsilon$ is a small value to avoid dividing by zero. The term $\text{Sim}_{ij}$ measures the similarity between the $i$th and $j$th paths that can be measured in a data-driven manner:
\begin{equation}\label{eq:sim}
\text{Sim}_{ij}=\frac{1}{K}\sum_{k=1}^{K}\sum\limits^W_{x=1}\sum^H\limits_{y=1}\min(S_{k_i}(x,y),S_{k_j}(x,y)),
\end{equation}
where $W$ and $H$ are the width and height of the input images, respectively. By resizing $S_{k_i}$ and $S_{k_j}$ to the input image resolution and normalizing them into probability distributions, $\text{Sim}_{ij}$ measures the average histogram interactions between the saliency maps estimated by two paths. By maximizing $\Omega_{r}$, the similarity between selected and unselected paths can become very high, leading to a less-redundant multi-path module.

The representativeness term $\Omega_r$ is defined between selected and unselected paths, while the diversity term $\Omega_d$ is defined only on the selected ones that aims to maximize their difference
\begin{equation}\label{eq:penaltyDiv}
\Omega_d = \frac{\sum\limits^M_{i=1}\sum\limits^M_{j=1}\delta{}(i\neq{}j)\cdot{}\alpha_i\cdot{}\alpha_j\cdot{}(1-\text{Sim}_{ij})}{\sum\limits^M_{i=1}\sum\limits^M_{j=1}\delta{}(i\neq{}j)\cdot{}\alpha_i\cdot{}\alpha_j+\epsilon}.
\end{equation}
We can see that this term will penalize the co-selection of two highly similar paths.

By incorporating \eqref{eq:penaltyRep} and \eqref{eq:penaltyDiv} into the optimization objective \eqref{eq:optobj}, we can obtain a binary optimization problem with quadratic terms. As $M$ is relatively small, enumeration can be adopted in ideal case. For large $M$, set optimization methods (\eg~submodular optimization) can quickly find a good local minimum.

\subsection{Spatial Prediction and Spatiotemporal Fusion}

\begin{figure}[t]
	\begin{center}
		\includegraphics[width=1.0\columnwidth]{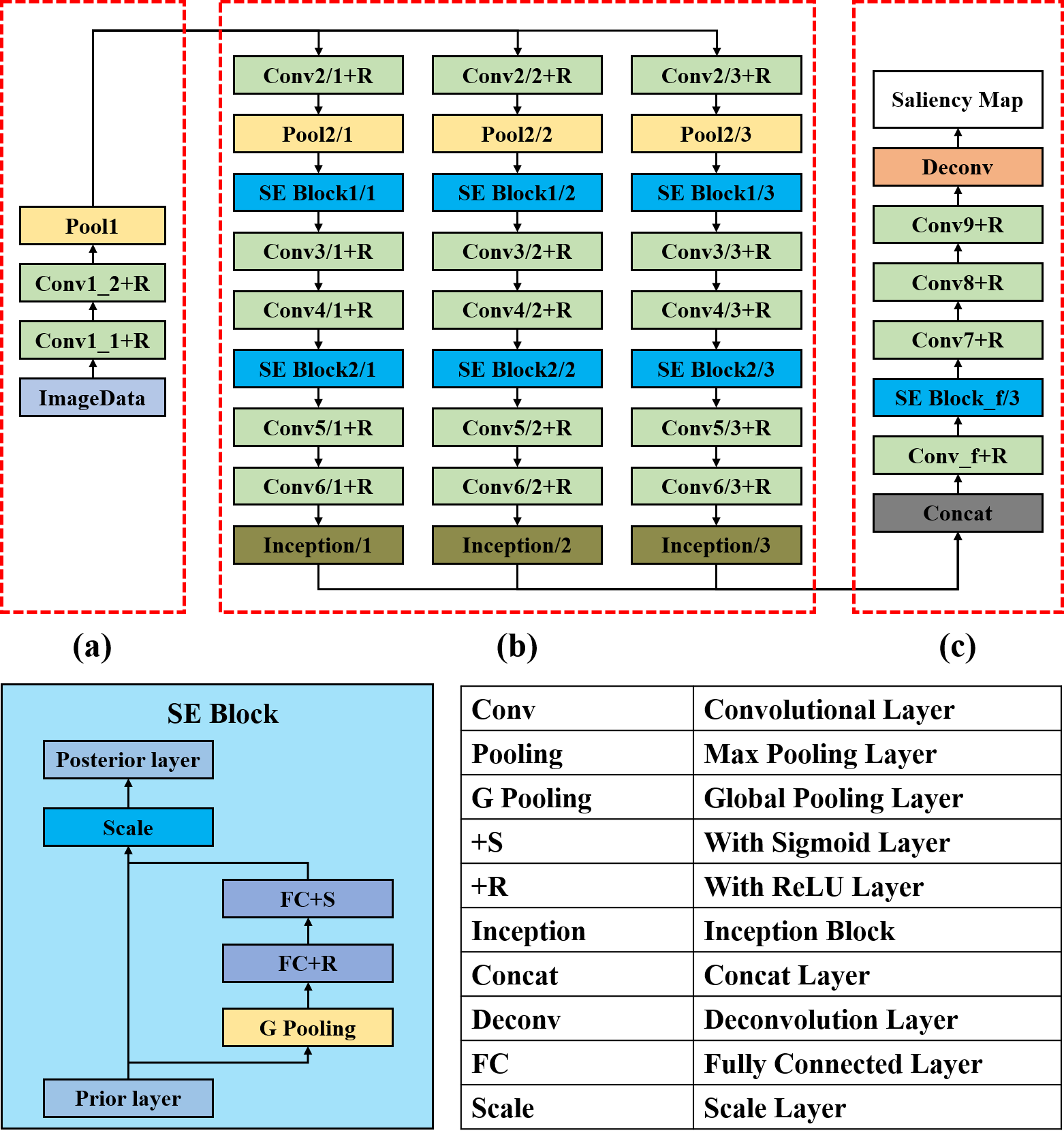}
	\end{center}
	\caption{The structure of MM-Net, including (a) low-level module, (b) multi-path module and (c) fusion module. }
	\label{fig:network structure}
\end{figure}

Since the regular GPU with 11G memory (\eg, GTX 1080Ti) can not afford the training of the proposed model with a large number of classic methods, we optimize the path settings within the acceptable range of hardware to seek a complexity-accuracy trade-off. By solving \eqref{eq:optobj} defined over the similarity matrix of ten paths, we obtain three representative paths (\ie, paths pre-trained by IT, QDCT, and SUN in the given parameters). After that, these selected paths are fused to output the spatial saliency maps. The overall structure of MM-Net can be found in  Fig.~\ref{fig:network structure}. In training MM-Net, the parameters of the fusion module are randomly initialized and then optimized with a learning rate of $5\times{}10^{-6}$.

Beyond the spatial fusion of paths, another necessary fusion is the spatial and temporal saliency maps. Let $S_k$ be the spatial saliency map given by MM-Net and $T_k$ be the temporal saliency map given by an existing temporal saliency model (\eg, \cite{fang2014video}). Inspired by \cite{liu2014superpixel}, we propose to spatiotemporally fuse $S_k$ and $T_k$ in adaptive fashion:
\begin{equation}
\label{eq:refineMap}
S_k^*=\lambda\cdot S_k^{int}+(1-\lambda)\cdot{}S_k^{sel},
\end{equation}
where $S_k^*$ is the refined saliency map, $S_k^{int}$ is the collaborative interaction of $S_k$ and $T_k$, and $S_k^{sel}$ is the selected spatial or temporal saliency map according to a heuristic rule. $\lambda$ is a positive scalar/weight to balance $S_k^{int}$ and $S_k^{sel}$. We first compute the spatial-to-temporal consistency score $C_{s2t}$ and temporal-to-spatial consistency score $C_{t2s}$:
\begin{equation}
C_{s2t} = e(S_k\odot{}T_k)/e(T_k),~ ~C_{t2s} = e(S_k\odot{}T_k)/e(S_k),
\end{equation}
where $e(\cdot)$ is the entropy function and $\odot{}$ indicates the per-pixel multiplication. We can see that the spatial-to-temporal consistency $C_{s2t}$ will be higher than the temporal-to-spatial consistency $C_{t2s}$ if the temporal saliency map is cleaner, and vice versa. As a result, the collaboration interaction map can be computed by emphasizing the cleaner map:
\begin{equation}\label{eq:intMap}
S_k^{int}=\frac{C_{t2s}\cdot{}T_k+C_{s2t}\cdot{}S_k}{C_{t2s}+C_{s2t}}.
\end{equation}

Let $d_k^{S}$ and $d_k^{T}$ be the weighted average of distances of salient pixels to their gravity centers in $S_k$ and $T_k$, respectively. The $S_k^{sel}$ is defined as the map with more compact salient regions:
\begin{equation}\label{eq:selMap}
S_k^{sel}=\left\{
\begin{aligned}
S_k &  & \text{if}~d_k^{S}\leq{}d_k^{T} \\
T_k &  & \text{otherwise}
\end{aligned}
\right.
\end{equation}
Intuitively, we can trust $S_k^{int}$ if the spatial and temporal saliency maps are highly consistent. If not, we can select the most compact map as the final prediction. Let $d_k^{int}$ be the average weighted distances of salient pixels to their gravity center in $S_k^{int}$, the parameter $\lambda$ can be computed as
\begin{equation} \label{eq:weight}
\lambda=\left\{
\begin{aligned}
\min(C_{t2s}&, C_{s2t}) &  & \text{if}~d_k^{int}<\omega\cdot\min(d_k^S,d_k^T) \\
&0 &  & \text{otherwise}
\end{aligned}
\right.
\end{equation}
where $\omega$ is a predefined weight that is empirically set to 2.1 (its influence will be discussed in experiments). By incorporating \eqref{eq:intMap}, \eqref{eq:selMap} and \eqref{eq:weight} into \eqref{eq:refineMap}, we can adaptively fuse the spatial and temporal saliency maps.

\section{Experiments} \label{sec:experiments}

We test our approach using the proposed aerial video saliency dataset AVS1K as well as the latest video saliency dataset DHF1K \cite{wang2018deep} that are fulfilled with daily scenarios captured by digital cameras and mobile phones. Both datasets, to the best of our knowledge, are currently the largest in their own domain. For DHF1K we use the official split, which contains 600, 100 and 300 videos for training, validation and testing, respectively.

On these two datasets, we compare MM-Net and two variants: MM-Net+ (with spatiotemporal refinement) and MM-Net- (without spatiotemporal refinement and the guidance of classic models). We also make comparisons with ten state-of-the-art models, including the heuristic group (H Group) HFT \cite{li2013visual}, SP \cite{li2014visual} and PNSP \cite{fang2014video}, the Non-Deep Learning Group (NL Group) SSD \cite{Li2015Finding} and LDS \cite{Fang2017Learning}, and the Deep Learning Group (DL Group) eDN \cite{Vig2014Large}, iSEEL \cite{tavakoli2017exploiting}, SalNet \cite{Pan2016Shallow}, DVA \cite{wang2018deep} and STS \cite{bak2017spatio}. Among these models, iSEEL and eDN are built on pre-extracted features and thus cannot be re-trained. For the other three models, we fine-tune them on the two datasets and use a mark $*$ to indicate the retrained model.

In the comparisons, we adopt five evaluation metrics, including the aforementioned AUC, sAUC and NSS as well as the Similarity Metric (SIM)~\cite{hou2013visual} and Correlation Coefficient (CC) \cite{borji2012boosting}. SIM is computed to measure the similarity of two saliency maps as probability distributions, while CC is computed as the linear correlation between the estimated and ground-truth saliency maps. Values of all the five metrics are positively correlated with the performance.

\subsection{Comparison with the State-of-the-art Models}
\begin{figure}[t]
	\begin{center}
		\includegraphics[width=1.0\columnwidth]{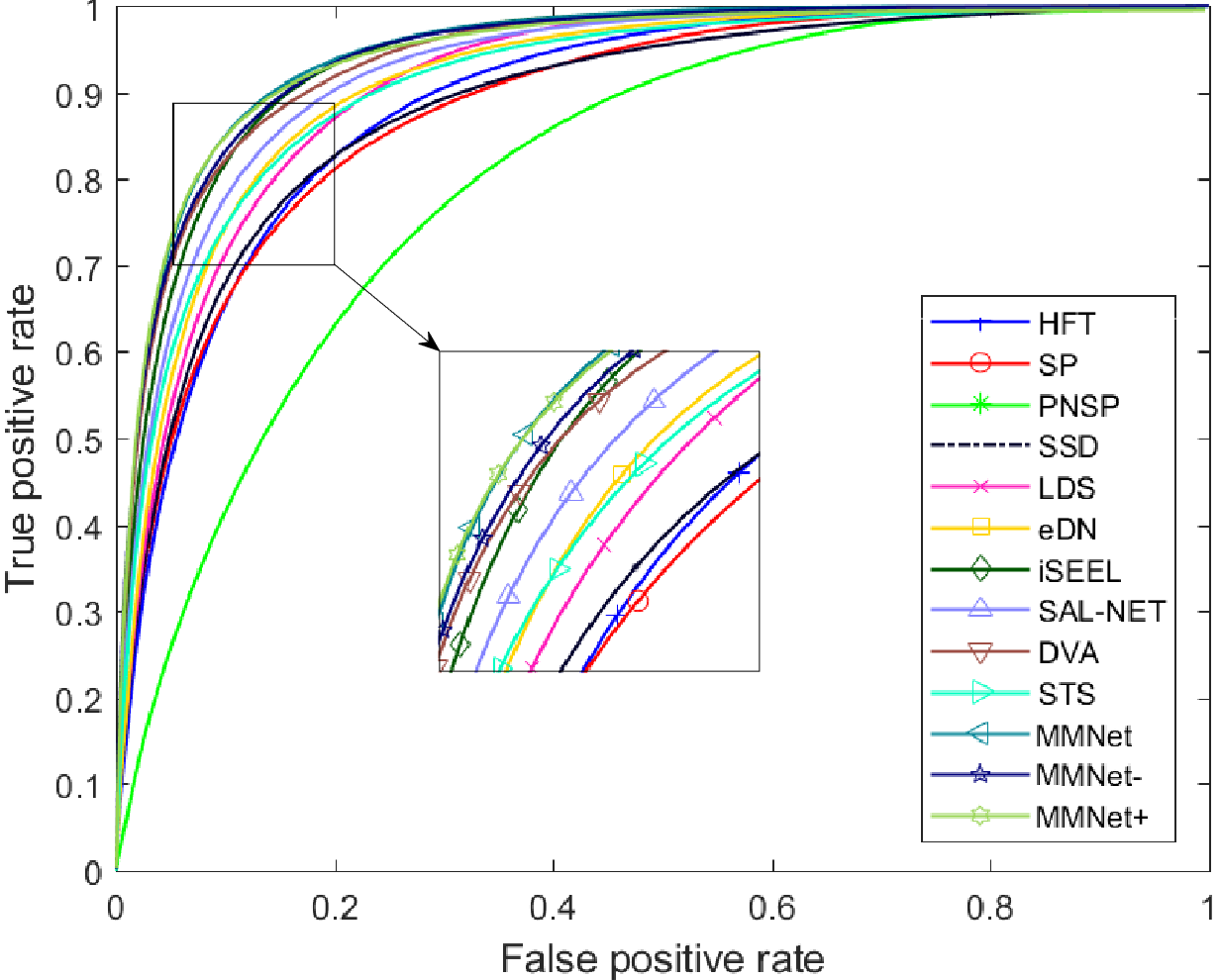}
	\end{center}
	\caption{ROC curves of 13 models on AVS1K.}
	\label{fig:ROC_AVS1K}
\end{figure}

\begin{figure*}[t]
	\begin{center}
		\includegraphics[width=0.98\textwidth]{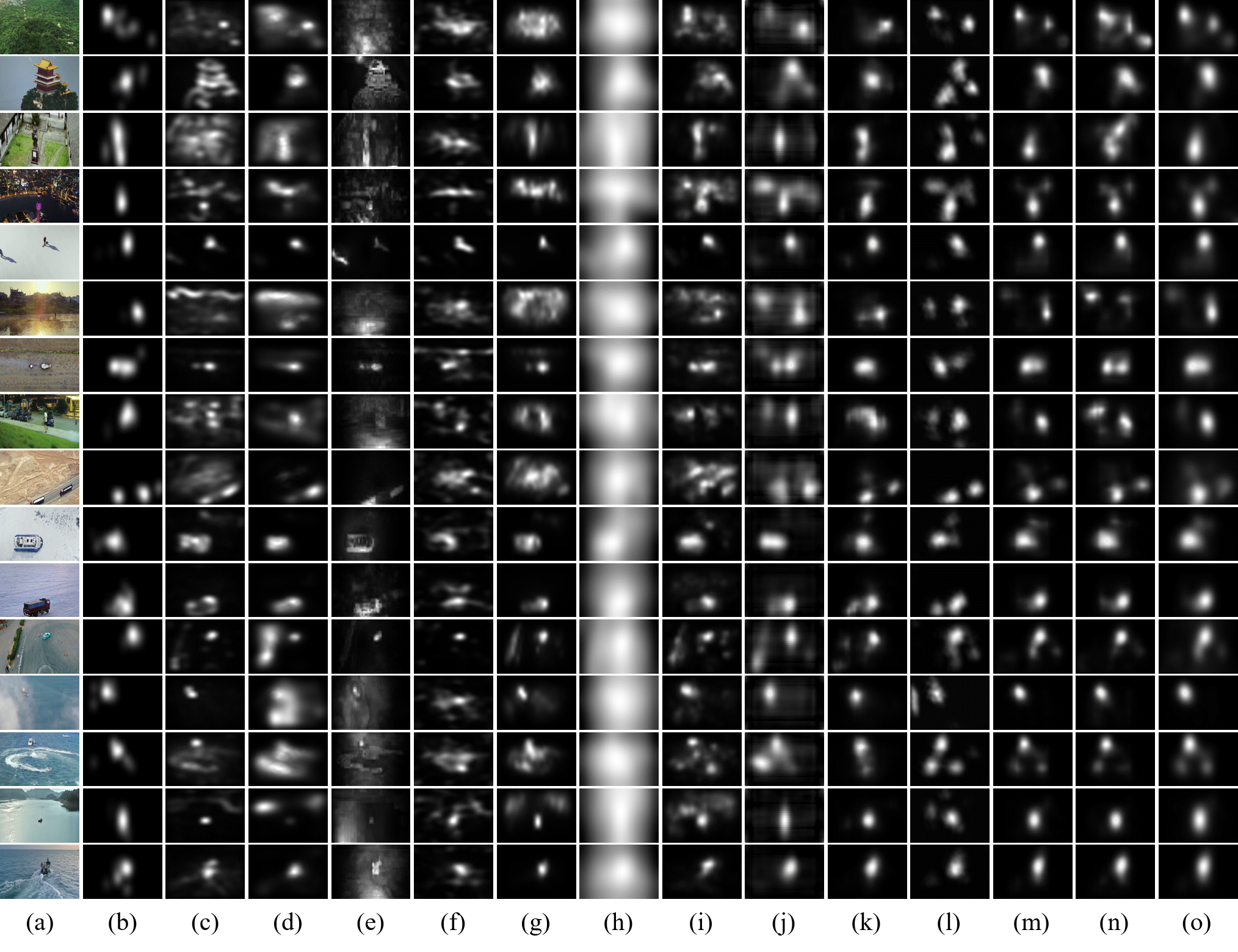}
	\end{center}
	\caption{Representative frames of state-of-the-art models on AVS1K. (a) Video frame, (b) Ground truth, (c) HFT, (d) SP, (e) PNSP, (f) SSD, (g) LDS, (h) eDN, (i) iSEEL, (j) SalNet, (k) DVA, (l) STS, (m) MM-Net, (n) MM-Net-, (o) MM-Net+.}
	\label{fig:Result_AVS1K}
\end{figure*}

Performance of 13 evaluated models on the AVS1K dataset are shown in \tabref{tab:performance_AVS1K}. We also illustrate the Receiver Operating characteristics Curves (ROC) in \figref{fig:ROC_AVS1K} and several representative results of these models in \figref{fig:Result_AVS1K}.

From \tabref{tab:performance_AVS1K}, we find that our fundamental multi-path network MM-Net-, in which the spatiotemporal refinement and the guidance of classic models are not used, still outperforms the other ten state-of-the-art models in terms of NSS and CC and ranks the second place in terms of AUC (worse than DVA), sAUC (worse than SalNet) and SIM (worse than DVA). Note that NSS is the primary metric recommended by many surveys on saliency evaluation metrics~\cite{wang2018deep, liu2016deep}. The impressive performance of MM-Net- can be explained by its multi-path structure. The low-level module of MM-Net- can extract many low-level preattentive features, based on which the multi-path module can further extract saliency cues at higher levels from different perspectives. These high-level saliency cues are then fused to obtain the final saliency map. In this way, MM-Net- has better representation capability when compared with traditional single path network (such as SalNet) and classic two-stream network for video (such as STS). Although DVA also adopts a multi-stream structure that directly fed supervisions into multi-layers, the MM-Net- still performs better than DVA in terms of sAUC, NSS and CC.

In \tabref{tab:performance_AVS1K}, we also find that MM-Net outperforms MM-Net- in terms of all metrics except AUC. This can be explained by the model guidance strategy adopted in training multiple paths. After initializing different network paths under the guidance of selected models with heuristically designed saliency features and rules, MM-Net inherently learns how to extract and fuse saliency-related features. In this manner, the biases of classic models can be further investigated and utilized to improve the saliency prediction accuracy. As a result, the effectiveness of the model guidance strategy can be well justified.

\begin{table}[t]
	\footnotesize
	\centering
	\caption{Benchmarking results on AVS1K. The best and runner-up models of each column are marked with bold and underline, respectively. Except our models, the other deep models fine-tuned on AVS1K are marked with *.}
	\label{tab:performance_AVS1K}
	\begin{tabular}{c|c|ccccc}
		\toprule
		\multicolumn{2}{c|}{Models}        &~AUC&~sAUC&~NSS&~SIM&~CC \tabularnewline
		\midrule
		\multirow{3}{*}{\bl{H}}&~HFT           &~0.789 &~0.715 &~1.671 &~0.408 &~0.539    \tabularnewline
		&~SP            &~0.781 &~0.706 &~1.602 &~0.422 &~0.520    \tabularnewline
		&~PNSP          &~0.787 &~0.634 &~1.140 &~0.321 &~0.370    \tabularnewline
		\midrule
		\multirow{2}{*}{\bl{NL}}&~SSD           &~0.737 &~0.692 &~1.564 &~0.404 &~0.503    \tabularnewline
		&~LDS           &~0.808 &~0.720 &~1.743 &~0.452 &~0.565    \tabularnewline
		\midrule
		\multirow{8}{*}{\bl{DL}}&~eDN           &~0.855 &~0.732 &~1.262 &~0.289 &~0.417    \tabularnewline
		&~iSEEL         &~0.801 &~0.767 &~1.974 &~0.458 &~0.636    \tabularnewline
		&~SalNet$^*$       &~0.797 &~0.769 &~1.835 &~0.410 &~0.593    \tabularnewline
		&~DVA$^*$           &~\underline{0.864} &~0.761 &~2.044 &~\underline{0.544} &~0.658    \tabularnewline
		&~STS$^*$           &~0.804 &~0.732 &~1.821 &~0.472 &~0.578    \tabularnewline
		&~MM-Net         &~0.858 &~\underline{0.771} &~\underline{2.110} &~\textbf{0.547} &~\underline{0.673}    \tabularnewline
		&~MM-Net-        &~0.860 &~0.768 &~2.087 &~0.541 &~0.666    \tabularnewline
		&~MM-Net+        &~\textbf{0.869} &~\textbf{0.784} &~\textbf{2.133} &~0.532 &~\textbf{0.682}    \tabularnewline
		\bottomrule
	\end{tabular}
\end{table}

From \tabref{tab:performance_AVS1K}, we also observe that MM-Net+ outperforms all the other models in terms of all metrics except SIM. This may be caused by the proposed spatiotemporal optimization algorithm. Based on the mutual consistency and weighted spatiotemporal saliency, the optimization algorithm tends to generate cleaner saliency maps with more compact salient regions (see \figref{fig:Result_AVS1K}). As a result, the most salient locations can pop-out in the saliency maps predicted by MM-Net+, leading to high NSS and sAUC scores.

We also find that the heuristic models in the H Group perform worse than the models in the NL Group and the DL Group. For H Group, the key issue here is that hand-crafted features designed for daily scenarios may be no longer suitable for the aerial scenarios. In other words, there may exist many irregular saliency visual patterns in aerial videos, which should be learned from data. This also explains the impressive performance of models in the DL Group since they can benefit from the powerful capabilities of CNNs in extracting hierarchical feature representations.

From these results, we conclude that, in aerial scenarios, the salient visual patterns as well as the feature fusion strategies may become remarkably different. As a result, it is necessary to learn the saliency cues and their fusion strategies that best characterize the salient visual patterns from the aerial perspective. In addition, there exist some inherent correlations between the daily and aerial scenarios, implying that a drone can also benefit from the knowledge encoded in previous models in learning how to look. By transferring such knowledge, a drone can gain a better capability of handling various visual patterns.

\subsection{Performance Analysis}
In this section, we conduct several experiments to analyze the performance of MM-Net+ (and MM-Net) from multiple perspectives, including parameter influences, generalization ability and performance on four subsets of AVS1K.

\begin{figure}[t]
	\begin{center}
		\includegraphics[width=1.0\columnwidth]{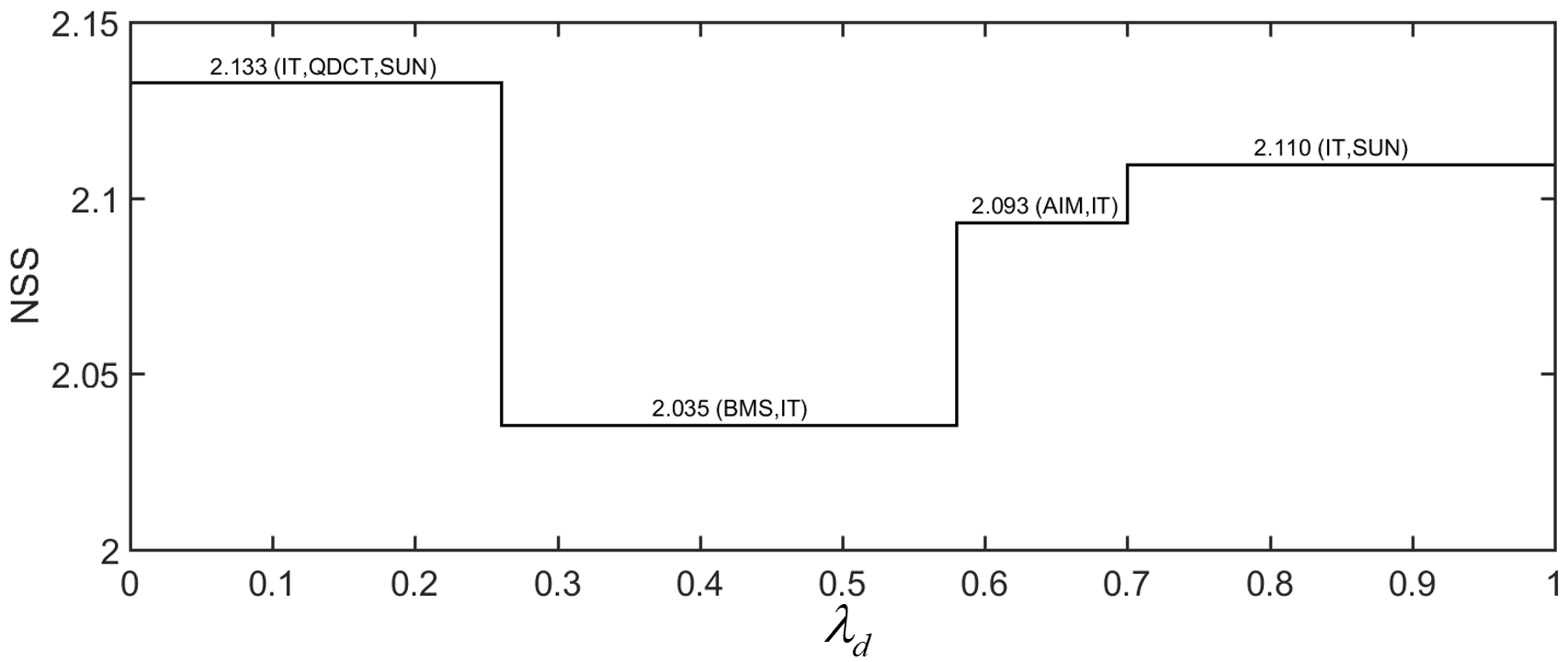}
	\end{center}
	\caption{Performance of MM-Net+ on AVS1K with the paths selected under different $\lambda_d$. In a wide range of $\lambda_d$, the selected paths stay almost stable (\ie, 2 or 3 paths)}
	\label{fig:lambda_d}
\end{figure}

In the first experiment, we analyze the parameter $\lambda_d$ in \eqref{eq:optobj} that is used to balance the representativeness and diversity in MM-Net+. The NSS curve of MM-Net+ on AVS1K with different $\lambda_d$ is shown in \figref{fig:lambda_d}. From \figref{fig:lambda_d}, we find that when $\lambda_d$ falls in [0, 0.24], MM-Net+ achieves the best performance (NSS=2.133) with three representative paths (IT, QDCT and SUN). When $\lambda_d$ grows, the number of selected path decreases. When $\lambda_d$ falls in [0.26, 0.56], MM-Net+ has lower complexity (only two selected paths, BMS and IT) as well as the lowest performance (NSS=2.035). When $\lambda_d$ grows larger, only two paths keep on being selected but they may be guided by two different models. For example, when $\lambda_d$ falls in [0.58, 0.68], MM-Net+ selects AIM and IT  with NSS=2.093. When $\lambda_d$ falls in [0.70, 1.00], MM-Net+ selects IT and SUN with NSS=2.093. To sum up, in a wide range of $\lambda_d$, the path selection algorithm tends to select two or three paths to reduce the model complexity. Therefore, we select $\lambda_d=0.2$ in all experiments for pursuing better performance at an acceptable performance.

\begin{figure}[t]
	\begin{center}
		\includegraphics[width=1.0\columnwidth]{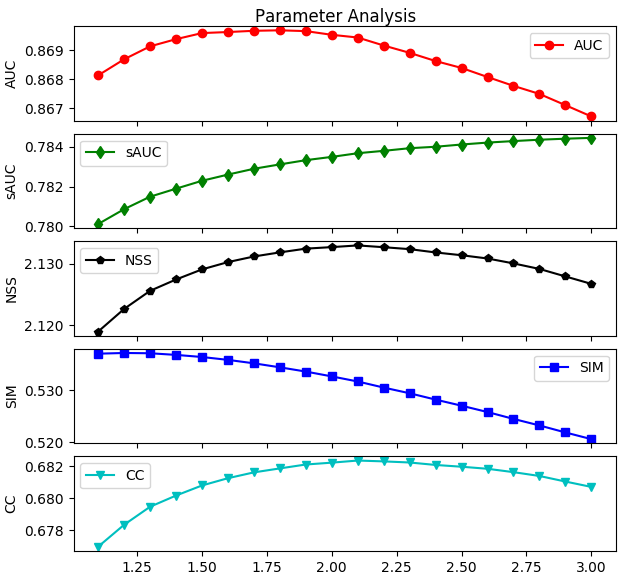}
	\end{center}
	\caption{Parameter analysis on AVS1K with different $\omega$ in the interval [1.1, 3.0].}
	\label{fig:parameter}
\end{figure}

In the second experiment, we analyze the parameter $\omega$ in \eqref{eq:weight} that serves as a threshold in computing $\lambda$ and further balance the fusion of spatial and temporal saliency maps in MM-Net+. The curves of AUC, sAUC, NSS, SIM and CC scores on AVS1K with different $\omega$ are shown in \figref{fig:parameter}.
We find that the AUC, NSS and CC curves are convex, the sAUC curve is monotonically increasing, and the SIM curve is monotonically decreasing. The overall performance is generally stable when $\omega$ falls between [1.8, 2.3]. With a small $\omega$, many saliency maps are refined with $\lambda=0$ (see \eqref{eq:weight}), implying that the spatial and temporal saliency maps cannot be adaptively fused in \eqref{eq:refineMap}. On the contrary, a large $\omega$ may generate non-zero $\lambda$ in most cases and thus lead to noisy saliency maps due to the additive fusion strategy of \eqref{eq:refineMap}. Therefore, we select $\omega=2.1$ in all experiments.

In the third experiment, we do some ablation studies to show how the path selection and spatiotemporal fusion affects or contributes to the saliency estimation. We present the performance of different settings of path selection in \tabref{tab:ablation}. From this table, we find that both the path selection and spatiotemporal fusion boost the performance of saliency prediction on aerial videos. The model with three paths is superior to those with two paths ones. This can be interpreted as the three paths that can encode more complete visual knowledge into the overall model, leading to more powerful intermediate representations. Meanwhile, we find that performance gaps exist between the different selection of classic models. The overall performance not necessarily positively relates to the performance of the classic models but depends on the fitness of the encoded knowledge.
\begin{table}[t]
\footnotesize
\centering
\begin{threeparttable}
\caption{Performance of different settings of path selection and spatiotemporal fusion on AVS1K. The best and runner-up models of each column are marked with bold and underline, respectively.}
\label{tab:ablation}
\setlength{\tabcolsep}{0.9mm}
\begin{tabular}{cccc|cccccc}
\toprule
Init & Fusion & Nums & Paths &~AUC&~sAUC&~NSS&~SIM&~CC  \tabularnewline
\midrule
                    &                   &3      &--                 &~0.860 &~0.768 &~2.087 &~\underline{0.541} &~0.666             \tabularnewline
 \checkmark         &                   &3      &IT,QDCT,SUN        &~0.858 &~0.771 &~2.110 &~\bl{0.547} &~0.673             \tabularnewline
 \checkmark         &\checkmark         &3      &IT,QDCT,SUN        &~\underline{0.869} &~\bl{0.784} &~\bl{2.133} &~0.532 &~\bl{0.682}             \tabularnewline
 \checkmark         &\checkmark         &2      &AIM,IT             &~0.867 &~0.778 &~2.093 &~0.527 &~0.670             \tabularnewline
 \checkmark         &\checkmark         &2      &BMS,IT             &~0.861 &~0.769 &~2.035 &~0.514 &~0.650             \tabularnewline
 \checkmark         &\checkmark         &2      &IT,SUN             &~\bl{0.872} &~\underline{0.781} &~\underline{2.110} &~0.528 &~\underline{0.674}             \tabularnewline
\bottomrule
\end{tabular}
\end{threeparttable}
\end{table}

In the fourth experiment, we compare MM-Net with 16 state-of-the-art models on the latest video saliency dataset DHF1K that are fulfilled with videos captured by digital cameras and mobile phones. The main objective of this experiment is to verify the generalization ability of MM-Net on various scenarios.
Quantitative results of these two models, after being fine-tuned on DHF1K, are shown in \tabref{tab:performance_DHF1K}. We can observe the proposed MM-Net outperforms the DVA on DHF1K in terms of all of metrics except for NSS. This proves the generalization ability of MM-Net, implying that the multi-path network architecture can be used for the saliency prediction task in both aerial and daily scenarios.

To verify the mutual generalization ability of saliency models in the aerial and general scenarios, we present the performance of two general models in \tabref{tab:generalization}. This table demonstrates the poor mutual generalization ability of saliency models in the aerial and general models.

Furthermore, to verify the performance of MM-Net+ on different scenarios, we show its performance on the four subsets of AVS1K in \tabref{tab:performance self}. MM-Net+ have relatively better performance on {AVS1K-H} and {AVS1K-V} than on {AVS1K-B} and {AVS1K-O}. In most aerial videos, humans and vehicles usually have relatively appropriate sizes and significant motions, which makes them easier to pop-out from the local context. On the contrary, buildings are static and usually have big sizes, making both the spatial and the temporal saliency prediction challenging. Similarly, {AVS1K-O} contains many diverse scenarios about planes, boats and animals. In these cases, the appearances and motion patterns of salient targets may change remarkably, making it difficult to separate them from distractors.

\begin{table}[t]
\footnotesize
\centering
\begin{threeparttable}
\caption{Performance comparison of 16 state-of-the-art models on DHF1K. The best and runner-up models of each column are marked with bold and underline, respectively.}
\label{tab:performance_DHF1K}
\begin{tabular}{l|l|ccccc}
\toprule
\multicolumn{2}{c|}{Models} &~AUC&~sAUC&~NSS&~SIM&~CC  \tabularnewline
\midrule
\multirow{6}{*}{\begin{sideways}\bl{Static models}\end{sideways}}
 &~ITTI~\cite{itti1998model}                &~0.774 &~0.553 &~1.207 &~0.162 &~0.233                                             \tabularnewline
 &~GBVS~\cite{harel2007graph}               &~0.828 &~0.554 &~1.474 &~0.186 &~0.283                                             \tabularnewline
 &~SALICON~\cite{huang2015salicon}          &~0.857 &~0.590 &~1.901 &~0.232 &~0.327                                 \tabularnewline
 &~Shallow-Net~\cite{Pan2016Shallow}        &~0.833 &~0.529 &~1.509 &~0.182 &~0.295                                             \tabularnewline
 &~Deep-Net~\cite{Pan2016Shallow}           &~0.855 &~0.592 &~1.775 &~0.201 &~0.331                                 \tabularnewline
 &~DVA~\cite{wang2018deep}                  &~\underline{0.860} &~\underline{0.595} &~\bl{2.013} &~\underline{0.262} &~\underline{0.358}                    \tabularnewline
\midrule
\multirow{10}{*}{\begin{sideways}\bl{Dynamic models}\end{sideways}}
 &~PQFT~\cite{guo2010novel}                 &~0.699 &~0.562 &~0.749 &~0.139 &~0.137                                             \tabularnewline
 &~Seo \etal~\cite{seo2009static}           &~0.635 &~0.499 &~0.334 &~0.142 &~0.070                                             \tabularnewline
 &~Rudoy \etal~\cite{rudoy2013learning}     &~0.769 &~0.501 &~1.498 &~0.214 &~0.285                                             \tabularnewline
 &~Hou \etal~\cite{hou2009dynamic}          &~0.726 &~0.545 &~0.847 &~0.167 &~0.150                                             \tabularnewline
 &~Fang \etal~\cite{fang2014video}          &~0.819 &~0.537 &~1.539 &~0.198 &~0.273                                             \tabularnewline
 &~OBDL~\cite{hossein2015many}              &~0.638 &~0.500 &~0.495 &~0.171 &~0.117                                             \tabularnewline
 &~AWS-D~\cite{leboran2016dynamic}          &~0.703 &~0.513 &~0.940 &~0.157 &~0.174                                             \tabularnewline
 &~OM-CNN~\cite{jiang2017predicting}        &~0.856 &~0.583 &~1.911 &~0.256 &~0.344         \tabularnewline
 &~Two-stream~\cite{bak2017spatio}          &~0.834 &~0.581 &~1.632 &~0.197 &~0.325                                             \tabularnewline
 &~\bl{Ours\tnote{*}}                       &~\bl{0.875} &~\bl{0.627} &~\underline{1.972} &~\bl{0.271} &~\bl{0.389}    \tabularnewline
\bottomrule
\end{tabular}
\begin{tablenotes}
    \footnotesize
    \item[*] Tested on the validation set of DHF1K.
\end{tablenotes}
\end{threeparttable}
\end{table}

\begin{table}[t]
\footnotesize
\centering
\begin{threeparttable}
\caption{Performance comparison of two general models on AVS1K and DHF1K.} \label{tab:generalization}
\setlength{\tabcolsep}{3.8mm}{
\begin{tabular}{l@{}|ccccc}
 \toprule
  ~Method&~AUC&~sAUC&~NSS&~SIM&~CC\tabularnewline \midrule
  ~DVA\tnote{*}                                  &~0.807 &~0.610 &~1.071 &~0.354 &~0.351    \tabularnewline
  ~SalNet\tnote{*}                               &~0.754 &~0.617 &~0.980 &~0.295 &~0.319    \tabularnewline
  ~DVA\tnote{+}                                  &~0.822 &~0.574 &~1.430 &~0.240 &~0.283    \tabularnewline
  ~SalNet\tnote{+}                               &~0.799 &~0.574 &~1.312 &~0.189 &~0.261    \tabularnewline
\bottomrule
\end{tabular}
}
\begin{tablenotes}
    \footnotesize
    \item[*] Trained on DHF1K and tested on AVS1K.
    \item[+] Trained on AVS1K and tested on the validation set of DHF1K.
\end{tablenotes}
\end{threeparttable}
\end{table}

\begin{table}[t]
	\centering{
		\caption{Performance of MM-Net+ on subsets of AVS1K. The best and runner-up models of each column are marked with bold and underline, respectively.}
		\label{tab:performance self}
        \setlength{\tabcolsep}{3.4mm}{
		\begin{tabular}{c|ccccc}
			\toprule
			Subset&~AUC&~sAUC&~NSS&~SIM&~CC \tabularnewline \midrule
			{AVS1K-B}&~\underline{0.872}&~0.780&~1.974&~0.543&~0.681 \tabularnewline
			{AVS1K-H}&~\bl{0.892}&~\underline{0.808}&~\underline{2.482}&~\underline{0.548}&~\underline{0.729} \tabularnewline
			{AVS1K-V}&~0.865&~\bl{0.826}&~\bl{2.497}&~\bl{0.566}&~\bl{0.759} \tabularnewline
			{AVS1K-O}&~0.859&~0.757&~1.926&~0.503&~0.630 \tabularnewline
			\bottomrule
		\end{tabular}
        }
	}
\end{table}

\section{Conclusion}
\label{sec:conclusion}
In this work, we introduce a large-scale video dataset for aerial saliency prediction. Based on this dataset, we propose MM-Net, a baseline model for aerial saliency prediction, which adopts a multi-path network structure and a model-guided training strategy to transfer human knowledge from classic models into the network paths. A spatiotemporal optimization algorithm is also proposed to fuse the spatial and temporal saliency maps.
Experimental results demonstrate the superior performance of our proposed models.

In the future work, we will explore the feasibility of learning a saliency model with few or none domain-specific training data. Such one-shot or zero-shot learning may further help the deployment of visual saliency models in many unknown scenarios.

\section*{Acknowledgment}

This work was supported in part by National Natural Science Foundation of China (No. 61922006, No. 61825101 and No. 61532003), and the Beijing Nova Program (No. Z181100006218063).

\ifCLASSOPTIONcaptionsoff
  \newpage
\fi

\bibliographystyle{IEEEtran}
\bibliography{MMNetBib}

%

\vspace{-1cm}

\begin{IEEEbiography}[{\includegraphics[width=1in,height=1.25in,clip,keepaspectratio]{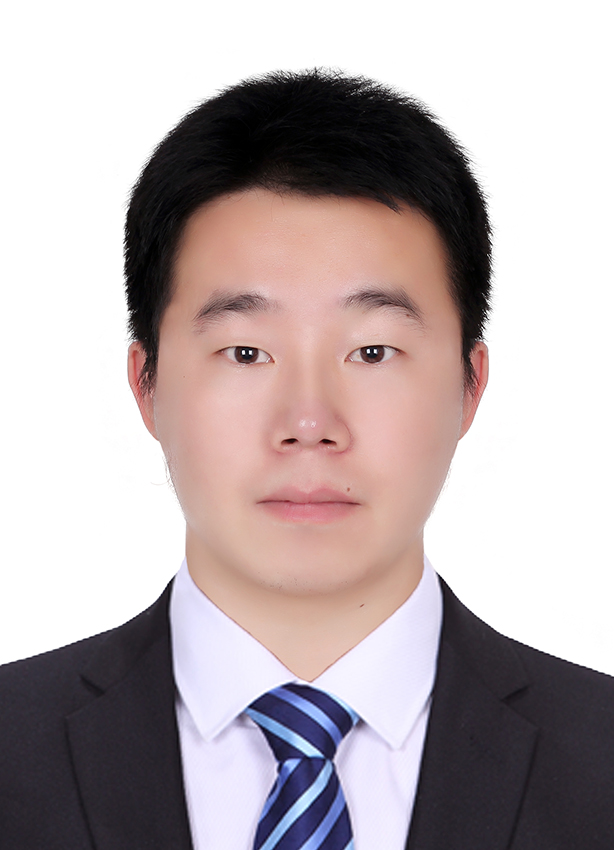}}]{Kui Fu} is currently pursuing the Ph.D. degree with the State Key Laboratory of Virtual Reality Technology and System, School of Computer Science and Engineering, Beihang University. His research interests include computer vision and image understanding.
\end{IEEEbiography}

\begin{IEEEbiography}[{\includegraphics[width=1in,height=1.25in,clip,keepaspectratio]{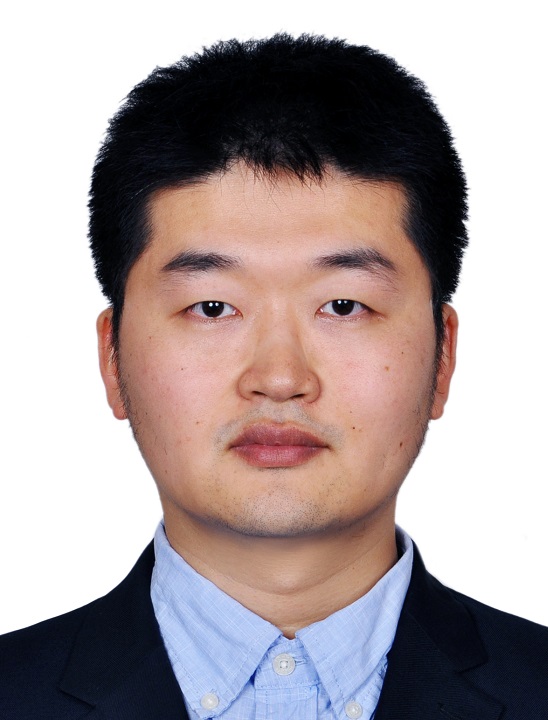}}]{Jia Li} (M'12-SM'15) received the B.E. degree from Tsinghua University in 2005 and the Ph.D. degree from the Institute of Computing Technology, Chinese Academy of Sciences, in 2011. He is currently a Full Professor with the School of Computer Science and Engineering, Beihang University, Beijing, China. Before he joined Beihang University in Jun. 2014, he used to conduct research in Nanyang Technological University, Peking University and Shanda Innovations. He is the author or coauthor of over 70 technical articles in refereed journals and conferences such as TPAMI, IJCV, TIP, CVPR and ICCV. His research interests include computer vision and multimedia big data, especially the understanding and generation of visual contents. He is supported by the Research Funds for Excellent Young Researchers from National Nature Science Foundation of China since 2019. He was also selected into the Beijing Nova Program (2017) and ever received the Second-grade Science Award of Chinese Institute of Electronics (2018), two Excellent Doctoral Thesis Award from Chinese Academy of Sciences (2012) and the Beijing Municipal Education Commission (2012), and the First-Grade Science-Technology Progress Award from Ministry of Education, China (2010). He is a senior member of IEEE, CIE and CCF. More information can be found at http://cvteam.net.
\end{IEEEbiography}

\begin{IEEEbiography}[{\includegraphics[width=1in,height=1.25in,clip,keepaspectratio]{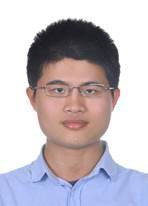}}]{Yu Zhang} is currently a researcher with SenseTime Research, Beijing China. He received his B.E. degree in 2012 and Ph.D. degree in 2018 from the School of Computer Science and Egnineering, Beihang University. His research interests include computer vision and image/video processing.
\end{IEEEbiography}

\begin{IEEEbiography}[{\includegraphics[width=1in,height=1.25in,clip,keepaspectratio]{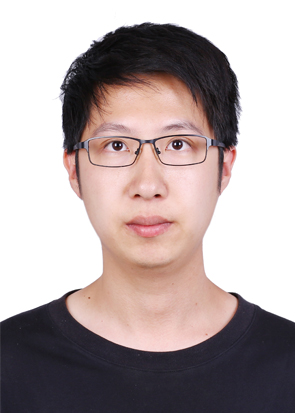}}]{Hongze Shen} received his Master degree in 2020 from the School of Computer Science and Engineering, Beihang University. His research interests include computer vision and image understanding.
\end{IEEEbiography}

\begin{IEEEbiography}[{\includegraphics[width=1in,height=1.25in,clip,keepaspectratio]{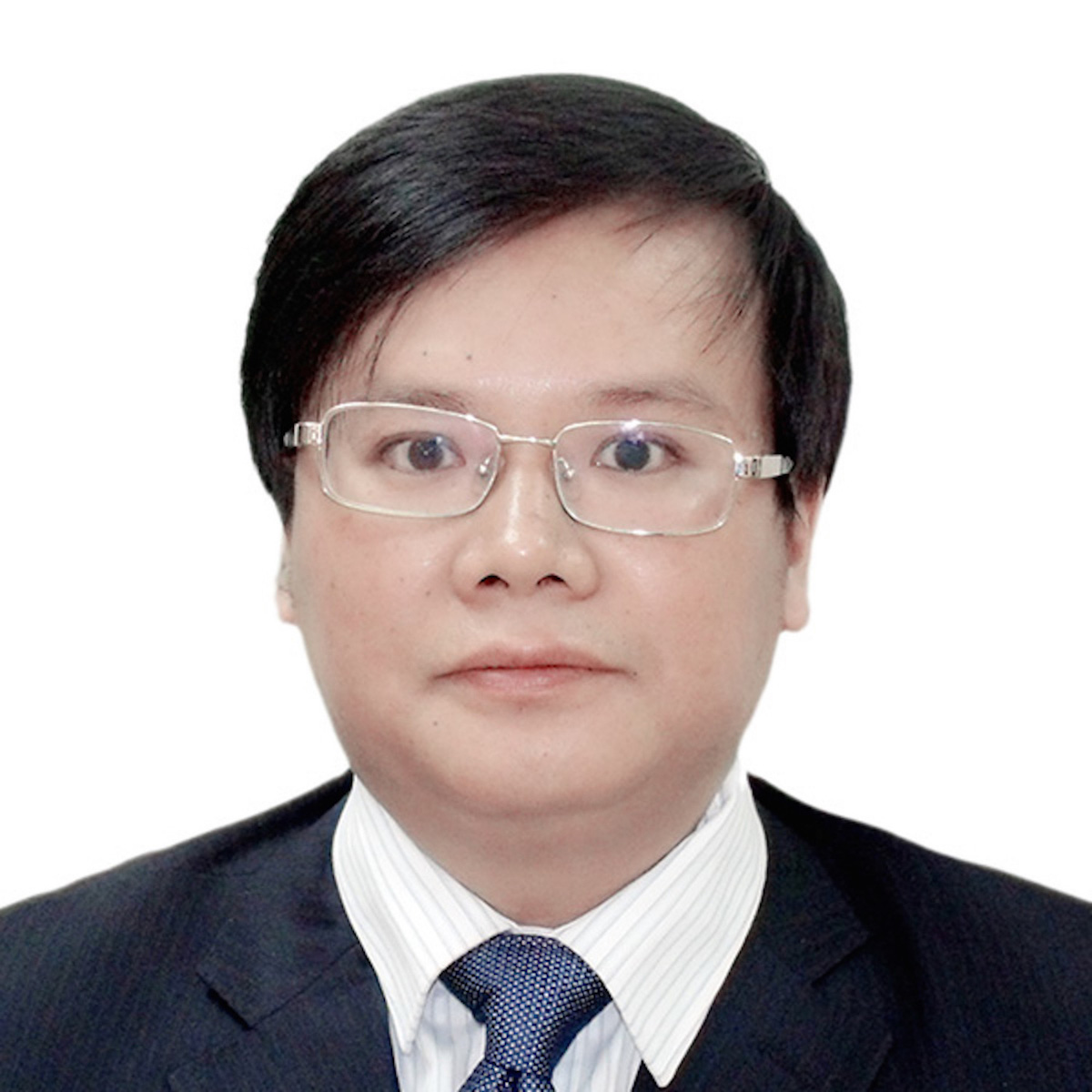}}]{Yonghong Tian}  (S'00-M'06-SM'10) is currently a Boya Distinguished Professor with the School of EECS, Peking University, China, and is also the deputy director of Artificial Intelligence Research Center, Peng Cheng Laboratory, Shenzhen, China. His research interests include computer vision, multimedia big data, and brain-inspired computation. He is the author or coauthor of over 180 technical articles in refereed journals such as IEEE TPAMI/TNNLS/TIP/TMM/TCSVT/TKDE/TPDS, ACM CSUR/TOIS/TOMM and conferences such as NeurIPS/CVPR/ICCV/AAAI/ACMMM/WWW.

Prof. Tian was/is an Associate Editor of IEEE TCSVT (2018.1-), IEEE TMM (2014.8-2018.8), IEEE Multimedia Mag. (2018.1-), and IEEE Access (2017.1-). He co-initiated IEEE Int'l Conf. on Multimedia Big Data (BigMM) and served as the TPC Co-chair of BigMM 2015, and aslo served as the Technical Program Co-chair of IEEE ICME 2015, IEEE ISM 2015 and IEEE MIPR 2018/2019, and General Co-chair of IEEE MIPR 2020. He is the steering member of IEEE ICME (2018-) and IEEE BigMM (2015-), and is a TPC Member of more than ten conferences such as CVPR, ICCV, ACM KDD, AAAI, ACM MM and ECCV. He was the recipient of the Chinese National Science Foundation for Distinguished Young Scholars in 2018, two National Science and Technology Awards and three ministerial-level awards in China, and obtained the 2015 EURASIP Best Paper Award for Journal on Image and Video Processing, and the best paper award of IEEE BigMM 2018. He is a senior member of IEEE, CIE and CCF, a member of ACM.
\end{IEEEbiography}




\end{document}